\documentclass{article}
\usepackage{spconf,amsmath,graphicx,hyperref}

\usepackage{booktabs}
\usepackage{algorithm}
\usepackage[switch]{lineno}
\usepackage{multirow}
\usepackage{subfig}
\usepackage{amsmath,amsfonts}
\usepackage{algorithmic}
\usepackage{graphicx}
\usepackage{textcomp}
\usepackage{xcolor}
\usepackage{comment}

\title{Learning Fill-in Reduction Ordering via Graph Policy Optimization for Sparse Matrices}
\name{Author(s) Name(s)\thanks{Thanks to XYZ agency for funding.}}
\address{Author Affiliation(s)}
\name{Ziwei Li$^{1,2}$, Shuzi Niu$^{1\ast}$, Huiyuan Li$^{1\ast}$, Tao Yuan$^{1\ast}$, and Wenjia Wu$^{1}$ \thanks{$\ast$ Corresponding authors: \{shuzi, huiyuan, yuantao\}@iscas.ac.cn}}
\address{%
$^{1}$ Institute of Software, Chinese Academy of Sciences, Beijing, China\\
$^{2}$ University of Chinese Academy of Sciences, Beijing, China}

\begin{document}
\ninept
\maketitle
\begin{abstract}

Matrix reordering in large sparse solvers seeks a permutation that minimizes factorization fill-in to reduce memory and computation. Because the minimum fill-in ordering problem is NP-complete and fill-in is implicit in the sparsity pattern, graph-theoretic heuristics are used. Existing reinforcement learning methods either ignore sparsity patterns--missing the global fill-in--or lack local exact fill-in feedback. We propose a graph policy optimization method, modeling fill-ins from global and local views: both the policy and value networks use a multi-hop graph neural backbone to embed global fill-in; the policy further interacts with symbolic factorization over graphs to extract local, step-level fill-ins, and the resulting feedback is aligned with the value network via an adaptive saturation function to improve convergence. On the SuiteSparse Matrix Collection, our method achieves mean reductions of 29.3\% in fill-ins and 31.3\% in peak memory usage over state-of-the-art baselines.
\end{abstract}
\begin{keywords}
Reinforcement Learning, Graph Neural Network, Matrix Reordering, Fill-in Reduction
\end{keywords}

\section{Introduction}

Matrix reordering is a key step that can substantially reduce memory usage when solving sparse linear systems. Many large-scale scientific simulations---such as structural engineering and computational fluid dynamics---lead to large sparse linear systems \(Ax=b\). A matrix \(A\in\mathbb{R}^{n\times n}\) is sparse when only a small fraction of its entries are nonzero~\cite{Wilkinson1971}, allowing storage and computation to focus on the nonzeros. In practice, one solves these systems by factoring \(A\) as \(A=LU\), where \(L\) is lower-triangular and \(U\) is upper-triangular. This LU factorization is Gaussian elimination in matrix form, realized via row operations. In sparse settings, the factors \(L\) and \(U\) often contain many more nonzeros than \(A\); these new entries are fill-ins, which drive memory and runtime.

\begin{figure}[tbp]
    \centering
    \includegraphics[width=1.05\linewidth]{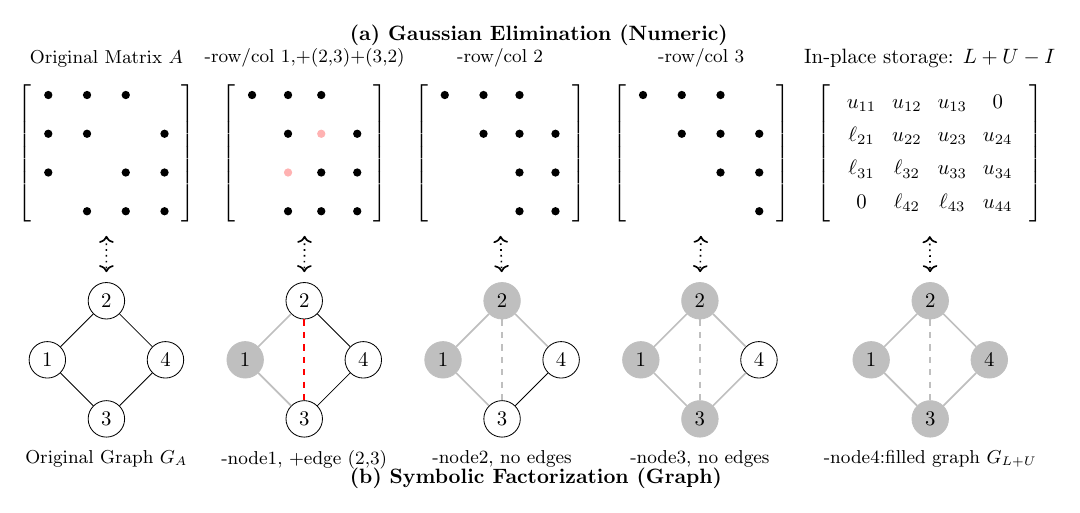}
    \vspace{-20pt}
    \caption{Step-level Fill-ins Generated from Symbolic Factorization}
    \label{fig:align}
\end{figure}

The Rose--Tarjan fill path theorem~\cite{Fill-Path} states that fill-ins are generated from the sparsity pattern of the matrix and the elimination ordering. The sparsity pattern of \(A\) is described by its adjacency graph \(G\) with a node for each row/column and an edge for each nonzero (undirected when \(A\) is symmetric). Practically, symbolic factorization is an efficient graphical algorithm that leverages this theorem to predict fill-ins and estimate the memory and computational requirements of the subsequent numerical factorization. An intuitive understanding is illustrated in Fig.~\ref{fig:align}, showing that eliminating a column in Gaussian elimination corresponds to removing a node and connecting its neighbors into a clique; the new nonzeros correspond to the added edges. Generally, matrix reordering is intrinsically a node reordering problem over the adjacency graph. 

However, computing the minimum fill-in ordering problem is NP-complete~\cite{Fill-Path,Yannakakis1981}. 
Graph-theoretic algorithms rely on heuristics such as node degrees and separators to deduce the node decision per step and finally produce approximate elimination orders. Approximate Minimum Degree (AMD)~\cite{rose1972graph,AMD,davis2004colamd} and Nested Dissection (ND)~\cite{ND} are representative degree- and separator-based methods, respectively. Such algorithms are usually complex and specially designed for certain kind of matrices.

Deep learning helps obtain a general reordering model. But the learning objective function, the number of fill-ins, is difficult to formalize explicitly and optimize directly. Deep reinforcement learning (DRL) instead minimizes the fill-in number iteratively via environment feedback. One related DRL method DRL\_ND~\cite{GP} uses a graph neural network on the adjacency graph as the node decision agent and learns agent parameters by minimizing node separators. It exploits the sparsity pattern, but was not designed straightly for fill-in minimization. Another related DRL method AlphaElim~\cite{alphaelim} applies a convolutional neural network over the matrix as the column elimination agent and trains agent parameters by minimizing fill-ins. It takes substantial computation for useless numeric operations and ignores the sparsity pattern. 
Therefore, we focus on leveraging the sparsity pattern to learn a general reordering agent model by directly optimizing the number of fill-ins.

To solve this problem, we propose Graph Policy Optimization (GPO) for sparse matrix reordering to reduce fill-in. 
First, we show that fill-ins are generated from the sparsity pattern of the matrix rather than specific numeric values. GPO operates on the adjacency graph of the sparsity pattern to model the entire fill-in generation process. It employs a policy network based on a graph neural network with multi-hop aggregation to choose which node to eliminate next. Crucially, the action is informed by feedback from a theoretically sound symbolic factorization process that calculates the resulting fill-ins. In addition, the value network refines its output via an adaptive saturation function to align with this feedback and stabilize the learning process.
We train on sparse matrix benchmarks generated via Delaunay triangulation~\cite{AMG,GP,alphaelim}, and evaluate on the SuiteSparse Matrix Collection~\cite{sparsematrix}. Metrics include mean fill-in count and mean peak memory usage during LU factorization, and GPO outperforms graph-theoretic and DRL baselines.

\section{Method}
\subsection{Fill-in Generation}
Given a sparse matrix $A$, fill-ins arise from matrix factorization, most commonly LU factorization~\cite{George1981Computer,Duff1986Direct,davis2016survey}. It is expressed as \(A=LU\) and derived from the Gaussian elimination process. \(L\) and \(U\) are the lower and upper triangular matrices respectively and \(L\) has a unit diagonal \(I\). For simplicity, we mainly consider the symmetric positive definite matrices \(A\in\mathbb{R}^{n\times n}\). Thus the Cholesky factor \(L\) of matrix \(A\) is a lower triangular matrix satisfying \(A=LL^T\), where \(L^T\) is the corresponding upper triangular factor \(U\).
The number of nonzero entries in \(A\) is denoted as \(\text{nnz}(A)\). Fill-ins are defined as new nonzero elements that are in \(L+U-I\) but not in \(A\). 

Fill-ins are completely embedded in sparsity patterns. An \(n\times n\) sparse matrix \(A\) induces an adjacency graph \(G_A=(V_A,E_A)\): each row/column corresponds to a node \(v\in V_A\), and each nonzero \(A_{ij}\neq 0\) corresponds to an edge \(e_{ij}\in E_A\). When \(A\) is symmetric, \(G_A\) is undirected. The Fill-Path Theorem~\cite{Fill-Path} states that an entry \(L_{ij}\) in the factor is nonzero iff there exists a path between \(i\) and \(j\) in \(G_A\) whose internal nodes are eliminated earlier than both \(i\) and \(j\). Given an elimination order, this theorem allows one to enumerate all potential fill-ins by tracing such connection paths. Symbolic factorization process in Alg.\ref{alg:graphcf} is an efficient and precise implementation of this basic idea to predict the fill-in pattern from the adjacency graph.

\begin{algorithm}[!htbp]
\caption{Basic Symbolic Cholesky Factorization Process}
\label{alg:graphcf}
\begin{algorithmic}[1]
\REQUIRE Graph $G_A $ and Elimination Ordering $\pi$
\ENSURE Fill Patterns $\mathcal{F}$,\\
Filled graph $G_L$ with the adjacency matrix $L+L^T$

\STATE $G_0 \gets G_A$
\STATE $\mathcal{F} \gets \{\emptyset\}_{k=0}^{n-1}$ 
\FOR{$k \gets 0$ \textbf{to} $n-1$}
  \STATE $v \gets \pi(k)$ 
  \STATE $N_{v}^{(k)} \gets \{u|(u,v)\in E_{k}\}$
  \FOR{ $u \in N_v^{(k)}$}
  \FOR{$w\in N_v^{(k)}-\{u\}$}
    \IF{$(u,w) \notin E_{k}$}
      \STATE $\mathcal{F}_k \gets \mathcal{F}_k \cup \{ (u,w) \}$ 
    \ENDIF
    \ENDFOR
  \ENDFOR
  \STATE $G_{k+1}\gets (V_{k} - \{v\}, (E_{k} - N_v^{(k)}\times \{v\})\cup\mathcal{F}_k)$
\ENDFOR

 \STATE $\mathcal{F}\gets\cup_{k=1}^n\mathcal{F}_k$
 \STATE $G_L\gets (V_0,E_0\cup\mathcal{F})$
\end{algorithmic}
\end{algorithm}

Given an elimination ordering \(\pi\) and the adjacency graph \(G_A\), fill-ins are predicted by sequentially eliminating nodes as Alg.\ref{alg:graphcf}. When the \(k\)-th node in the ordering, \(v=\pi(k),0\leq k\leq n-1\), is eliminated from the current graph \(G_k=\bigl(V_k,\,E_k\bigr)\), namely elimination graph, edges connected to \(v\) will be removed, denoted by \(N_v^{(k)}\times \{v\}\), where \(N_v^{(k)}\) is \(v\)'s neighbor set~\cite{liu1990role}. At the same time, additional edges $\mathcal{F}_k$ will be added among node pairs in \(N_v^{(k)}\), where there are no edges in the current graph \(G_k\). Each fill edge \((u,w)\), \(u,w\in N_v^{(k)}\) naturally satisfies the condition that the elimination orders of \(u\) and \(w\) are greater than \(k\). In addition to $\mathcal{F}_k$, the elimination graph will be updated as \(G_{k+1}\) according to Line 13 of Alg.\ref{alg:graphcf} and the subgraph of \(N_v^{(k)}\) will become a clique. We repeat this elimination process \(n\) times until all nodes have been removed.
Ultimately, the fill-ins generated throughout the entire elimination process can be expressed as \(\cup_{k=0}^{n-1}\mathcal{F}_k\).
For a specific graph, different orderings lead to different fill-ins. In this sense, matrix reordering is to find an elimination ordering \(\pi\) over the adjacency graph \(G_A\) to minimize the number of generated fill-ins \(|\cup_{k=0}^{n-1}\mathcal{F}_k|\).

\begin{figure}[tbp]
\centering
\includegraphics[width=0.8\linewidth]{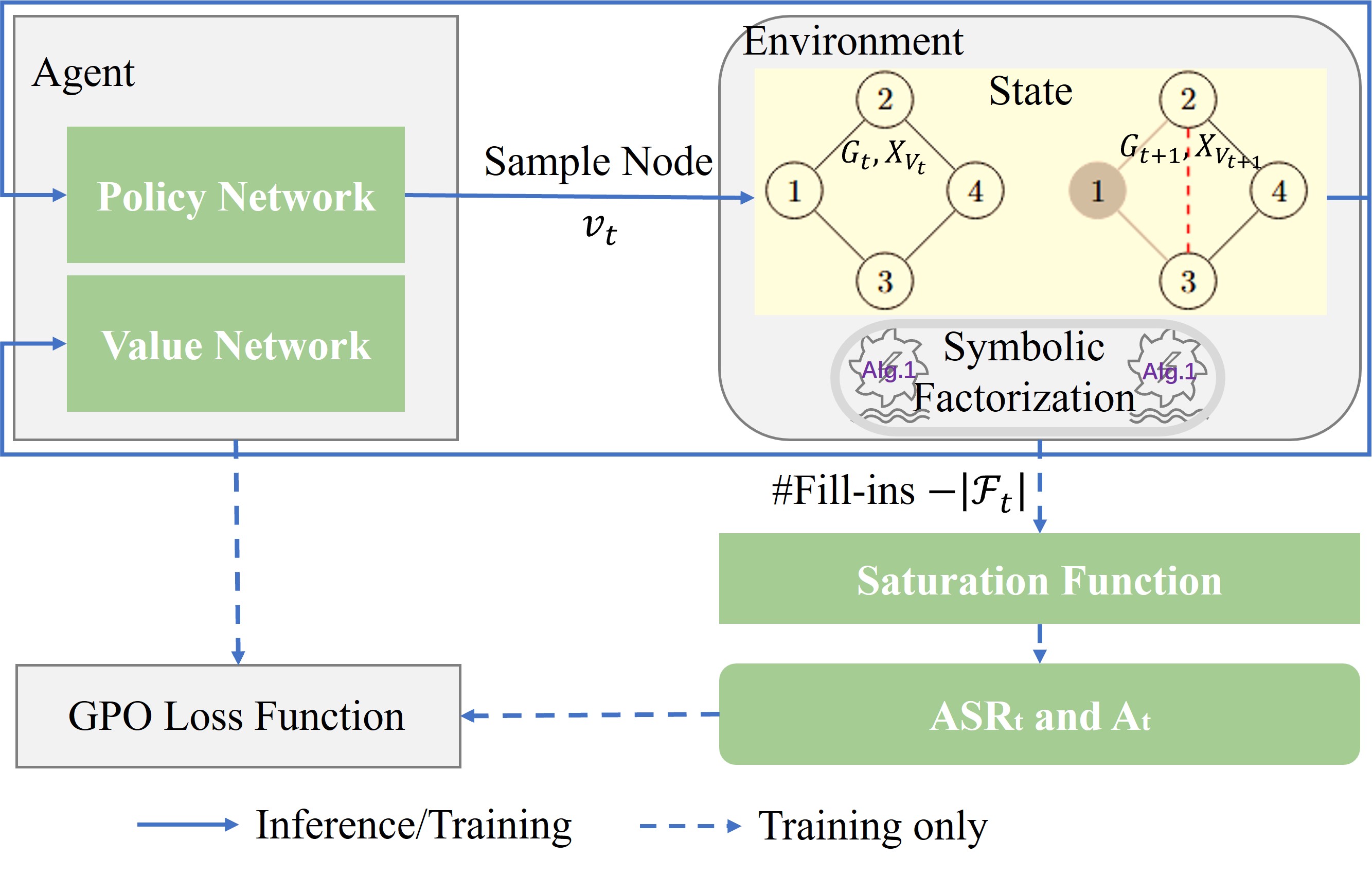}
\vspace{-10pt}
\caption{Architecture of the Graph Policy Optimization Framework}
\label{fig:RL}
\end{figure}

\subsection{Graph Policy Optimization Framework}

We propose a Graph Policy Optimization (GPO) framework to learn effective node elimination policies that minimize fill-ins in Fig.~\ref{fig:RL}. With reinforcement learning, GPO addresses the non-differentiability of the fill-in objective. Fill-ins are exactly collected from the interaction between GNN agent and the symbolic factorization process on the adjacency graph per node elimination step. The policy and value network parameters with MixHopConv backbones are learned by minimizing the adaptive saturation version of fill-ins.

We adopt a reinforcement learning formulation with the following components.
\textbf{The environment} is the symbolic factorization process illustrated in Alg.~\ref{alg:graphcf}.
\textbf{The state} at time step \(t\) is the elimination graph \(G_t=(V_t,E_t)\) together with its node features \(X_{V_t}\). The node features are the node degree in \(G_t\), \(\deg(v)\), and collective influence~\cite{CI} in Eq.(\ref{eq:ci}).
\(
\mathrm{CI}(v)=(\deg(v)-1)\!\!\sum_{u\in N(v)}(\deg(u)-1).
\)
\textbf{The state space} comprises all elimination graphs and their associated node features that can be generated under different node-elimination orders.
\textbf{The action space} consists of all nodes \(V_A\) in the original graph \(G_A\); at time step \(t\), the action is chosen from the nodes \(V_t\) in the current elimination graph \(G_t\).
\begin{equation}
\footnotesize
    \mathrm{CI}(v)=(\deg(v)-1)\!\!\sum_{u\in N(v)}(\deg(u)-1)
    \label{eq:ci}
\end{equation}

\begin{figure}[!htbp]
\centering
\includegraphics[width=0.5\linewidth]{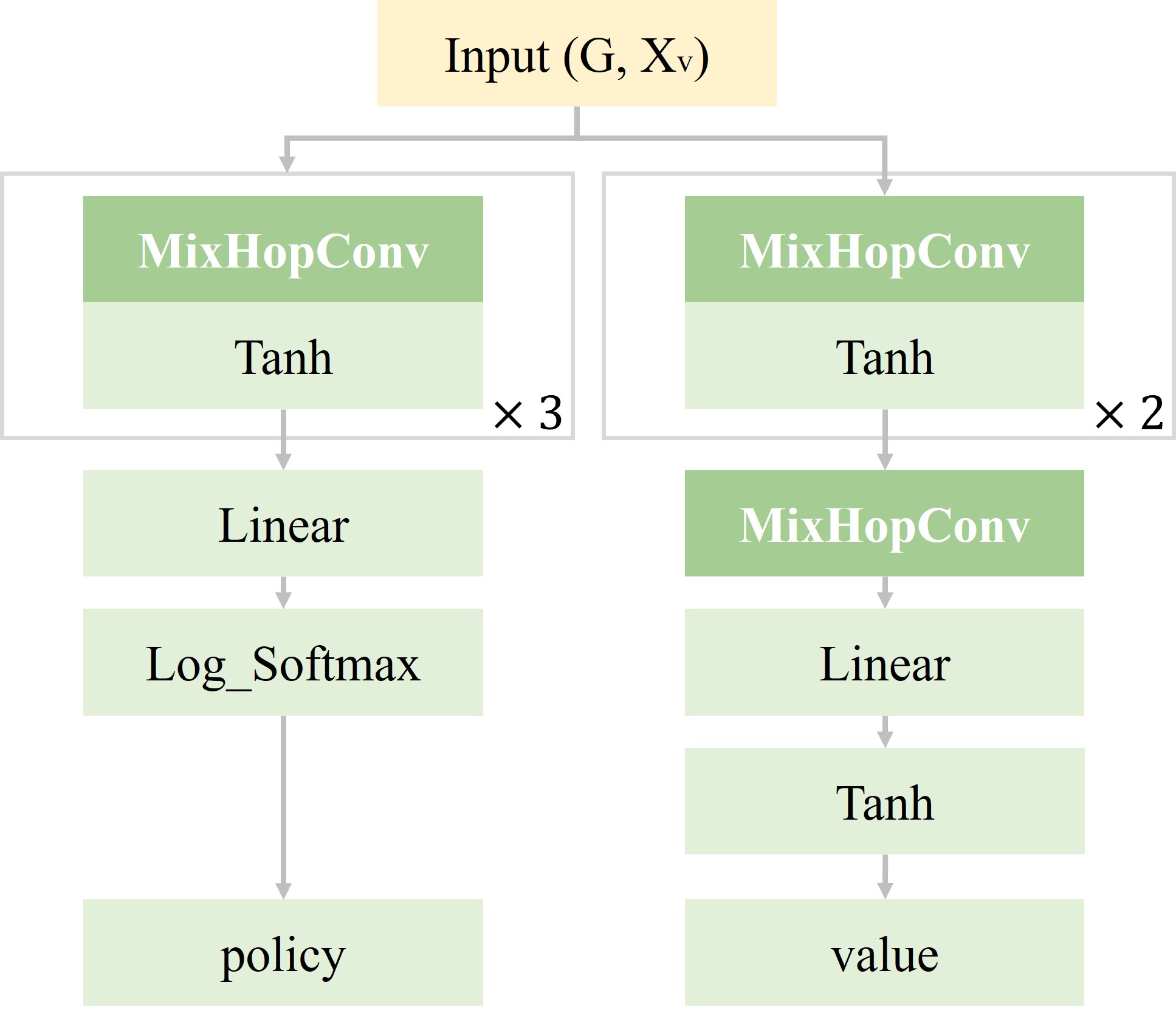}
\vspace{-8pt}
\caption{Network Architecture of GNN-Agent}
\label{fig:network}
\end{figure}

\textbf{The agent} consists of a policy and value network illustrated in Fig.~\ref{fig:network}. The policy network takes the elimination graph \(G_t\) and the node-feature matrix \(X_{V_t}\), and outputs a log-probabilities \(\pi(V_t\mid G_t)\) over candidate nodes for elimination. The value network computing a scalar \(\mathrm{Value}(G_t)\) as the expected average adaptive saturation return achievable from \(G_t\). Both networks are composed of several MixHopConv layers~\cite{mixedhop} with tanh activation functions to extract meaningful node embeddings.
The key difference lies in the final transformation layer. For the actor network, the output from the last tanh activation is passed through a linear transformation followed by a log\_softmax layer, producing the probability distribution for node selection. In contrast, for the critic network, a linear transformation is applied before the final tanh activation to generate a representation of the graph, which is used to evaluate the current state.
At each time step, the agent selects a node to eliminate according to \(\pi(V_t\mid G_t)\) and performs the elimination as in Alg.~\ref{alg:graphcf}, which yields a \textbf{state transition} \(G_t\to G_{t+1}\). 
\begin{equation}
\footnotesize
    \mathrm{ASR}_t(|E_t|,R_t)=\frac{|E_t|+R_t}{|E_t|-R_t}
    \label{eq:asr}
\end{equation}

\textbf{The reward} \(r_t\) is defined as the negative count of added edges, \(r_t=-|\mathcal{F}_t|\). To reduce variance and make training more stable, the \textbf{Adaptive Saturation Return} normalizes the number of fill-ins as Eq.(\ref{eq:asr}).
where \(|E_t|\) is the number of edges in \(G_t\), and \(-R_t\) is the number of newly added edges from time \(t\) to termination, i.e., \(R_t=\sum_{k=t}^{n-1} r_k\).
\textbf{The advantage function} is defined as the difference between the adaptive saturation return and the value function, \(A_t=\mathrm{ASR}_t-\mathrm{Value}(G_t)\); it measures how much better the selected action is relative to the expected return.
The actor loss \( L_a \) is defined as the negative sum of log probabilities multiplied by their respective advantages in Eq.(\ref{eq:gpoloss}).
The critic loss \( L_c \) is designed as the mean squared error between the adaptive saturation returns and the state value estimates in Eq.(\ref{eq:gpoloss}).
GPO training process is in Alg.~\ref{GPO Training Process}.
\begin{equation}
\footnotesize
L_a = -\frac{1}{n} \sum_{t=0}^{n-1} \pi(v_t|G_t) \cdot A_t, L_c = \frac{1}{n} \sum_{t=0}^{n-1} A_t^2
    \label{eq:gpoloss}
\end{equation}

\begin{algorithm}[!htbp]
\caption{Graph Policy Optimization Training Process}
\label{GPO Training Process}
\begin{algorithmic}[1]
\REQUIRE Training Graph Set $\{G\}$

\FOR{each graph $G$ in $\{G\}$}
    \STATE Initialize $G_0$ = $G$, $X_V$
    \STATE Initialize policies $\mathcal{P}=[]$, values $\mathcal{V}=[]$
    \STATE Initialize rewards $\mathcal{R}=[]$, the number of edges $\mathcal{E}=[]$

    \FOR{$t$ from $0$ to $n-1$}
        \STATE $\pi(V_t|G_t)$, $Value(G_t)$ $\leftarrow$ GNN($G_t$, $X_{V_t}$)
        \STATE Sample node $v_t$ from policy distribution $\pi(V_t|G_t)$ 
        
        \STATE $\mathcal{P}$.append($\pi(v_t|G_t)$), $\mathcal{V}$.append($Value(G_t)$)
        
        \STATE // eliminate the node, calculate fill-ins and $X_{V_t}$
        \STATE $G_{t+1}$, $\mathcal{F}_t$, $X_{V_{t+1}}$ $\leftarrow$ NodeElimination($G_t$, $v_t$)
        \STATE $\mathcal{R}$.append($-|\mathcal{F}_t|$), $\mathcal{E}$.append($|E_t|$)

    \ENDFOR

    \STATE Compute the Adaptive Saturation Return $ASR(\mathcal{E},\mathcal{R})$
    \STATE Compute the advantages $\mathcal{A} = ASR - \mathcal{V}$
    \STATE Compute the loss per episode $L_{actor}$ and $L_{critic}$
    \STATE UpdateGNNParameters($L_{actor},L_{critic}$)
\ENDFOR
\end{algorithmic}
\end{algorithm}

The whole updating step for each training graph \(G\) is from Line 2 to 16 of Alg.~\ref{GPO Training Process}. It includes initialization from Line 2 to 4, inference process along with fill-in collection from Line 5 to 12, and gradient backpropagation from Line 13 to 16. The initial state of the environment is represented as \( G_0 = G \) together with its node features \(X_V\). Both the agent inference process and fill-in collection process are interleaved within Line 5\(\sim\)12. For a new test graph, its complete elimination ordering is derived from \(\mathcal{P}\) after running Line 5\(\sim\)12.

\section{Experiments}
\subsection{Experimental Settings}

As training data, we generate 10{,}000 symmetric matrices via Delaunay triangulation~\cite{AMG,alphaelim,GP} with sizes \(n\in[60,200]\). For evaluation, we randomly sample from each category of the SuiteSparse Matrix Collection to form a 332-matrix test set spanning five domains: Structural Problem (SP), 2D/3D Problem (2D/3D), Circuit Simulation Problem (CSP), Electromagnetics Problem (EP), and Computational Fluid Dynamics Problem (CFDP). Sizes are grouped as S (10--500), M (500--1k), L (1k--10k), and XL (10k--100k), where "size" denotes the number of rows/columns; unsymmetric matrices are evaluated on the symmetrized form \(A+A^{\top}\). The test matrix distribution is in Table~\ref{test set}. Each MixHopConv layer is of 16 hidden units. We implement GPO with PyTorch and Adam optimizer with learning rate \(0.01\) for epoch 1 and \(0.001\) thereafter. Training phase is completed in 50 hours on an NVIDIA A100.

\begin{table}[tbp]
\caption{Test Matrix Count Distribution by Matrix Type and Size}
\label{test set}
\centering
\scriptsize
\resizebox{\columnwidth}{!}{
\begin{tabular}{lccccc}
\toprule
Data set &S(10-500) & M(500-1k) & L(1k-10k) &XL(10k-100k) & Total\\
\midrule
SP & 24&  20& 30 & 10 & 84\\
2D/3D & 14& 14& 10 & 10 & 48\\
CSP  & 6& —& 19 &9 & 34\\
EP   &8& 7 & 14 &6  & 35\\
CFDP  & 22& 23& 70 &16 & 131\\
\bottomrule
\end{tabular}
}
\end{table}

\begin{table*}[tbp]
\caption{Fill-in Ratio of Sparse Matrix Reordering Algorithms on SuiteSparse Test Set}
\label{tab:nnz}
    \centering
    \tiny
    \resizebox{\textwidth}{!}{
    \begin{tabular}{c|p{0.4cm}p{0.4cm}p{0.4cm}p{0.5cm}|p{0.4cm}p{0.4cm}p{0.4cm}p{0.4cm}|p{0.4cm}p{0.1cm}p{0.4cm}p{0.4cm}|p{0.4cm}p{0.4cm}p{0.4cm}p{0.4cm}|p{0.4cm}p{0.4cm}p{0.4cm}p{0.4cm}}
    \hline
     \multirow{2}{*}{Algorithm}  	&\multicolumn{4}{c|}{2D/3D Problem}				&\multicolumn{4}{c|}{Structural Problem}				&\multicolumn{4}{c|}{Circuit Simulation Problem}				&\multicolumn{4}{c|}{Computational Fluid Dynamics}				&\multicolumn{4}{c}{Electromagnetics Problem}\\
        \cline{2-21}
	&S	&M	&L	&XL	&S	&M	&L	&XL &S	&M	&L	&XL &S	&M	&L	&XL &S	&M	&L	&XL\\
 \hline
NATURAL&	5.5391 	&21.4712 	&37.6738 	&131.201 	&5.3414 	&11.0633 	&26.6641 	&53.151 	&10.5513 	&-	&200.05 	&723.78 	&3.7866 	&9.1709 	&15.3802 	&166.45 	&8.1252 	&23.4617 	&26.3767 	&369.265 \\
COLAMD	&1.8089 	&4.0584 	&5.3278 	&29.6666 	&2.0060 	&2.9777 	&5.0403 	&20.761 	&1.0625 	&-	&30.995 	&44.387 	&1.7982 	&4.1578 	&5.6925 	&16.243 	&0.8596 	&3.3734 	&4.8695 	&60.6559 \\
METIS	&1.3055 	&3.0115 	&\textbf{3.2885} 	&\textbf{11.6896} 	&1.3021 	&2.3826 	&3.1239 	&9.6555 	&0.5628 	&-	&6.4704 	&3.3489 	&1.2606 	&2.7999 	&3.2052 	&10.011 	&0.9679 	&3.0492 	&3.4340 	&21.3862 \\
DRL\_ND	&1.8362 	&5.1762 	&4.0304 	&14.6448 	&1.3647 	&2.6048 	&5.7978 	&13.237 	&0.7011 	&-	&6.9841 	&4.7575	&1.5166 	&3.5259 	&4.7636 	&25.670 	&1.1079 	&4.2830 	&7.4880 	&27.7424 \\
PPO	&1.6532 	&4.6778 	&3.4141 	&12.0328 	&2.3061 	&5.1168 	&3.3386 	&10.052 	&0.5438 	&-	&6.3444 	&3.7386	&2.0351 	&4.7473 	&3.4806 	&10.445 	&1.2033 	&3.6963 	&3.6507 	&21.6282 \\
GPO	&\textbf{1.1063} 	&\textbf{2.6354} 	&3.3353 	&11.9360 	&\textbf{1.2269} 	&\textbf{2.3053} 	&\textbf{3.0499} 	&\textbf{9.4997} 	&\textbf{0.4866} 	&-	&\textbf{6.2848} 	&\textbf{3.0067} 	&\textbf{0.9872} 	&\textbf{2.3636} 	&\textbf{3.1227} 	&\textbf{9.9480} 	&\textbf{0.7630} 	&\textbf{2.6253} 	&\textbf{3.4176} 	&\textbf{20.9969} \\
\hline
    \end{tabular}
    }
    
\end{table*}

\begin{table*}[tbp]
\caption{Nonzero Number nnz(L+U) and Peak Memory Usage in Kilobytes (PM\_KB) of LU Factorization with Different Reordering Methods}
\label{tab:lu_pk}
\centering
\tiny
\resizebox{\textwidth}{!}{
\begin{tabular}{l|r|rr|rr|rr|rr|rr|rr}
\hline
\multirow{2}{*}{Matrix} &
\multirow{2}{*}{Size(n)} &
\multicolumn{2}{c|}{NATURAL} &
\multicolumn{2}{c|}{COLAMD}  &
\multicolumn{2}{c|}{METIS}   &
\multicolumn{2}{c|}{DRL\_ND}   &
\multicolumn{2}{c|}{PPO}   &
\multicolumn{2}{c}{GPO}      \\
\cline{3-14}
& & nnz(L+U) & PM\_KB & nnz(L+U) & PM\_KB & nnz(L+U) & PM\_KB & nnz(L+U) & PM\_KB & nnz(L+U) & PM\_KB & nnz(L+U) & PM\_KB \\
\hline
Pres\_Poisson        & 14,822 & 10,229,633   &   273,888  &   5,504,778  &   183,176   &   4,996,318  &   158,736   &  5,896,452 & 186,148  & 5,259,803  & 165,832  & \textbf{4,735,200}   & \textbf{154,176} \\
bcsstk31             & 35,588 & 46,538,357   & 1,102,056  &  23,055,939  &   577,872   &  10,506,668  &   291,004   & 13,527,311 & 360,360  & 10,046,738 & 279,132  & \textbf{9,777,728}   & \textbf{273,484} \\
bcsstk36             & 23,052 & 20,624,794   &   528,192  &  14,112,629  &   383,308   &  10,933,097  &   306,324   & 10,536,260 & 296,024  &  8,796,149 & 256,880  & \textbf{7,674,468}   & \textbf{235,144} \\
coupled              & 11,341 & 40,547,108   & 1,008,928  &   1,067,616  &    65,824   &   1,066,795  &    63,684   &     59,412 & 479,395  & 1,019,026  & 62,548   & \textbf{426,787}     & \textbf{48,672}  \\
pli                  & 22,695 & 40,322,762   &   977,908  &  57,276,921  & 1,313,628   &  38,350,240  &   895,148   & 54,505,246 & 1,245,588& 38,599,143 & 901,276  &  \textbf{35,082,796}  & \textbf{819,808} \\
poisson3Da           & 13,514 & 166,114,198  & 3,653,376  &  15,177,132  &   378,388   &   5,617,200  &   167,004   &   6161,232 &  187,800 & 5,643,660  & 167,184  & \textbf{5,374,060}   & \textbf{161,768} \\
\hline
\end{tabular}
}
\end{table*}

\subsection{Evaluation Protocols}

We compare GPO with Natural Ordering and two representative classical graph-theoretic baselines---COLAMD~\cite{davis2004colamd} and ND (implemented via METIS~\cite{METIS} and referred to as METIS)——plus three RL baselines: (i) \emph{PPO}, obtained by replacing GPO's A2C-like loss with a PPO-style loss under identical training data and hyperparameters; (ii) \emph{DRL\_ND}~\cite{GP}, which learns graph partitions to minimize normalized node separators and then applies graph-theoretic reorderings within each subgraph (trained exactly as in~\cite{GP}); and (iii) \emph{AlphaElim}~\cite{alphaelim}, which treats the matrix as an image and combines CNNs with MCTS to align the reordering process with the numerical factorization steps of Gaussian elimination. To improve parallelism on large problems, we partition \(G_A\) using Nested Dissection and run GPO on each subgraph when the matrix size exceeds \(10^3\).

For each test matrix \(A\), we compute an ordering with each method, apply it to obtain \(A'\), and factorize \(A'\) using \texttt{splu}, a Python interface for \texttt{SuperLU}~\cite{superLU}, to produce \(L\) and \(U\). We report the fill-in ratio
\(
\mathrm{FIR}=\frac{\mathrm{nnz}(L+U-I)-\mathrm{nnz}(A)}{\mathrm{nnz}(A)},
\)
which normalizes the extra nonzeros by the original sparsity so matrices of different sizes and densities are comparable. We also report the peak memory during factorization—the maximum resident set size in kilobytes (PM\_KB)—capturing the footprint of the factors and all work arrays.
\subsection{Memory Usage Analysis of LU Factorization}
For each of the 19 matrix groups in Table~\ref{test set}, Table~\ref{tab:nnz} shows the average fill-in ratio for each reordering method, excluding AlphaElim. Since the source code of AlphaElim~\cite{alphaelim} is unavailable, it is evaluated separately at the end of this subsection. To better illustrate the impact of reordering on memory usage during LU factorization, we also report peak memory usage (PM\_KB) for a subset of matrices with \(n>10{,}000\), randomly sampled from the matrices listed in Table~\ref{test set}. Results are shown in Table~\ref{tab:lu_pk}.

As shown in Table~\ref{tab:lu_pk}, PM\_KB decreases as the fill-in drops, indicating lower peak working set during factorization. In Tables~\ref{tab:nnz} and~\ref{tab:lu_pk}, all reordering methods outperform Natural Ordering in both FIR and PM\_KB, underscoring the value of reordering. GPO consistently achieves the lowest FIR and the lowest PM\_KB across problem domains, attributable to alignment with symbolic factorization, MixHopConv GNN as backbone and proposed adaptive saturation return. One exception is that GPO exhibits a slightly higher fill-in ratio than METIS on the large and extra-large 2D/3D matrices. The reason lies in that the graph partitioning method has more effect on the fill-in ratio for larger 2D/3D matrices.

Reinforcement-learning methods achieve state-of-the-art results versus graph-theoretic baselines on most matrices in both metrics, highlighting the suitability of RL-based reorderings. Although training uses only Delaunay-triangulation matrices while testing uses real SuiteSparse matrices with markedly different sizes and sparsity patterns, the lower FIR and PM\_KB on the test set indicate strong generalization, especially for GPO. Moreover, despite being trained on small matrices (\(n\in[60,200]\)), we do not partition test matrices with \(n\le 1{,}000\) (S/M categories) and reorder them directly; even so, GPO attains the lowest FIR, demonstrating scalability.

We restrict the comparison to matrices evaluated by AlphaElim (Table~\ref{tab:alphaelim}). FIR(AlphaElim) is computed from its reported nonzero counts. From table, GPO performs comparably to AlphaElim on small matrices with \(n<1,000\), and even outperforms it on \texttt{olm500}. For large-scale matrices with sizes approaching one million, GPO demonstrates significantly better fill-in reduction.
However, the result on matrix \texttt{m\_t1} appears unusual: AlphaElim yields a negative fill-in ratio, which is counterintuitive considering the unstructured numerical values of the matrix. In all, results indicate that GPO achieves superior performance compared to AlphaElim, particularly on large-scale sparse matrices.
\begin{table}[tbp]
 \caption{Fill-in Ratio Comparison between AlphaElim and GPO}
\label{tab:alphaelim}
    \centering
    \scriptsize
    \begin{tabular}{c|ccc}
    \hline
	Matrix & Size ($n$) & FIR(AlphaElim) & FIR(GPO) \\
 \hline
mbeause   & 496     & \textbf{1.3101}   & 1.3123 \\
olm500    & 500     & 0.5381   & \textbf{0.0000} \\
ex27      & 974     & \textbf{0.6984}   & 0.7165 \\
m\_t1     & 97578   & -0.2415  & 5.4822 \\
tx2010    & 914231  & 290.7237 & \textbf{3.4145} \\
Hardesty1 & 938905  & 39.3514  & \textbf{16.7099} \\
\hline
    \end{tabular}
\end{table}

Overall, compared with state-of-the-art baselines (excluding Natural Ordering and AlphaElim), GPO achieves a \textbf{29.3\%} mean reduction in fill-in ratio. For matrices with \(n>10{,}000\), it further reduces PM\_KB by an average of \textbf{31.13\%}. These reductions translate into a smaller working set during LU factorization, which is critical for large-scale sparse computations.

\subsection{Ablation Study}
\begin{table}[tbp]
\caption{Ablation Study in Terms of Fill-in Ratio on 2D3D Problem}
\label{tab:ablation}
    \centering
    \resizebox{\columnwidth}{!}{
    \begin{tabular}{c|ccc|ccc}
    \hline
        	&DRL\_SF 	&DRL\_ND &$\uparrow$  &GPO	&GPO\_SAGE&$\uparrow$\\
         \hline
2D3D-S	 		&\textbf{1.2306} &1.8362    &-0.6056  &\textbf{1.1063} 	&1.1147 &-0.0084\\ 
2D3D-M	 	&6.2428  	&\textbf{5.1762}  &1.0666  &\textbf{2.6354} 	&5.5880 &-2.9526\\
2D3D-L	 	&\textbf{3.3438} 	&4.0304   &-0.6866  &\textbf{3.3353} 	 &3.3370 &-0.0017\\
2D3D-XL	 	&\textbf{11.9553} 	&14.6448  &-2.6895  &\textbf{11.9360} 	 &11.9370 &-0.0010\\
\hline
    \end{tabular}
    }  
\end{table}


Here we study the roles of two key components in GPO including MixHopConv backbone and the symbolic factorization process. To explore the role of agent backbone, we replace MixHopConv with the SAGEConv~\cite{sageconv} in GPO denoted as GPO\_SAGE. 
To probe into the reinforcement learning process, we keep the agent backbone in DRL\_ND, change its alignment with nested dissection into alignment with \textbf{S}ymbolic \textbf{F}actorization, which we refer to as DRL\_SF.

Comparing GPO and GPO\_SAGE in Table~\ref{tab:ablation}, results indicate that the MixHopConv backbone has a positive effect on the reordering method. Global context modeling in MixHopConv is helpful especially for large sparse matrices. 
DRL\_SF consistently achieves a lower fill-in ratio than DRL\_ND, except for medium-sized sparse matrices.
As the matrix size increases, the fill-in ratio discrepancy increases. It suggests the excellent effect of alignment with symbolic factorization on larger sparse matrices. Both components are essential for fill-in reduction of GPO.

\section{Conclusion}

Sparse matrix reordering seeks an elimination ordering that minimizes fill-ins during matrix factorization. To avoid costly numerical factorization and leverage graph structures, we analyze the correspondence between symbolic and numerical factorization and establish that fill-in depends on the original graph and the elimination order. To address the lack of an explicit fill-in formulation while capturing both global graph context and local fill-in pattern, we propose a reinforcement learning–based Graph Policy Optimization framework. It models symbolic factorization as environment dynamics, using the negative number of fill-ins as the reward. MixHopConv GNNs serve as actor–critic backbones to alleviate locality limitations. Experiments on real-world sparse matrices show that our method outperforms graph-theoretic and reinforcement learning baselines in both fill-in ratio and peak memory usage.

\section{Acknowledgments}
This research was supported by the National Key R\&D Program of China under Grant No. 2021YFB0300203.

\bibliographystyle{IEEEbib}
\bibliography{strings,refs}

\end{document}